\documentclass[letterpaper, 10 pt, conference]{ieeeconf}  

\IEEEoverridecommandlockouts                              

\usepackage{amsmath} 
\usepackage{amssymb}  
\usepackage{amsfonts}
\usepackage{fixmath}
\usepackage{graphicx}
\usepackage{color}
\usepackage{bm}

\title{\LARGE \bf
Nth Order Analytical Time Derivatives of Inverse Dynamics in Recursive and Closed Forms
}

\author{Shivesh Kumar$^{1}$ and Andreas M\"uller$^{2}$
\thanks{*Both authors have equal contribution.}
\thanks{$^{1}$Shivesh Kumar is with Robotics Innovation Center, DFKI GmbH, 28359 Bremen, Germany
        {\tt\small shivesh.kumar@dfki.de}}%
\thanks{$^{2}$Andreas Mueller is with the Institute of Robotics, Johannes Kepler University, 4040 Linz, Austria
        {\tt\small a.mueller@jku.at}}%
}

\begin{document}

\maketitle
\thispagestyle{empty}
\pagestyle{empty}

\begin{abstract}
Derivatives of equations of motion describing the rigid body dynamics are becoming increasingly relevant for the robotics community and find many applications in design and control of robotic systems. Controlling robots, and multibody systems comprising elastic components in particular, not only requires smooth trajectories but also the time derivatives of the control forces/torques, hence of the equations of motion (EOM). This paper presents novel $n^{\text{th}}$ order time derivatives of the EOM in both closed and recursive forms. While the former provides a direct insight into the structure of these derivatives, the latter leads to their highly efficient implementation for large degree of freedom robotic system.
\end{abstract}


\section{Introduction}

Rigid body dynamics algorithms and their derivatives find numerous
applications in the design optimization and control of modern robotic
systems. The equations of motion can be differentiated with respect to state
variables, control output (generalized forces), time and physical parameters
of the robot (see~\cite{Ott2013} for an overview). These derivatives can be
computed by several methods: 1) approximation by finite differences, 2)
automatic differentiation i.e. by applying the chain rule formula in an
automatic way knowing the derivatives of basic functions (cos, sin or exp),
3) closed form symbolic derivatives of the (Lagrangian) equation of motion
(EOM), and 4) recursive formulations exploiting the $O\left( n\right) $
algorithms to evaluate the EOM. While the first two methods are generic and
numerical in nature, the latter two are analytical in nature and exploit the
structure of the EOM.

Owing to the generality of automatic differentiation, it has been adopted by
two popular optimization based control frameworks namely Drake~\cite{drake}
and Control Toolbox~\cite{ETHControlToolbox}. The authors in~\cite%
{AutoDiffETHZ} argue that second-order or higher-order derivatives of the
equations of motion result in very large expressions and hence find it
questionable whether they should be implemented manually. However, there
have been attempts in the literature~\cite{Park2005},~\cite{Mansard2018} to
derive analytical and recursive partial first order derivatives of the rigid
body dynamics with respect to states and generalized forces. Recently, the
authors in~\cite{Mansard2018} demonstrated that it is worthy to investigate
efficient recursive formulations for the partial derivatives and presented
computational efficiency superior to that of automatic differentiation
without having to deal with costly technological setup of code generation.
These derivatives are useful in optimal control of legged robots (e.g.
differential dynamic programming in Crocoddyl framework~\cite{Mansard2019})
and their computational design \& optimization~\cite{Yamane2017}.

Time derivatives of the EOM are required for the control of robots with
compliant joints/gears. In particular, motion planning with higher-order
continuity \cite{ReiterTII2018} and flatness-based control of robots
equipped with series elastic actuators (SEA), respectively variable
stiffness actuators (VSA) \cite%
{deLuca1998,GattringerMUBO2014,PalliMelchiorriDeLuca2008} necessitate the
first and second time derivatives of the equations of motion (EOM) of the
robot. Therefore, recursive $O\left( n\right) $-algorithms for the
evaluation of the first and second time derivatives were developed \cite%
{BuondonnaDeLuca2015,BuondonnaDeLuca2016,Guarino2006,Guarino2009,ICRA2017}
extending existing $O\left( n\right) $-formulations for the evaluation of
EOM. The aim of this paper is to present novel $n^{\text{th}}$ order time
derivatives of equations of motion in both recursive and closed forms which
can be implemented easily in any rigid body dynamics library so that they
can provide time derivatives of motion equations until any order. Known
applications of this work include dynamically consistent smooth motion
planning with higher-order continuity~\cite{ReiterTII2018}, control of
robots with flexible joints. For the subsequent treatment, the EOM are
written in the form suitable for solving the inverse dynamics problem%
\begin{equation}
\mathbf{Q}=\mathbf{M}\left( \mathbf{q}\right) \ddot{\mathbf{q}}+\mathbf{C}%
\left( \dot{\mathbf{q}},\mathbf{q}\right) \dot{\mathbf{q}}+\mathbf{Q}_{%
\mathrm{grav}}\left( \mathbf{q}\right)  \label{EOM1}
\end{equation}%
where the vector of generalized coordinates $\mathbf{q}=\left( q_{1},\ldots
,q_{n}\right) ^{T}$ comprises the $n$ joint variables, $\mathbf{M}$ and $%
\mathbf{C}$ is the generalized mass and Coriolis matrix, respectively, and $%
\mathbf{Q}_{\mathrm{grav}}$ represents generalized gravity forces. Finally, $%
\mathbf{Q}$ are the generalized forces (drive forces/torques) required for a
prescribed motion $\mathbf{q}\left( t\right) $.

Throughout the paper, we will make repeated use of the Leibniz rule for the $%
n^{\text{th}}$ order derivative of a product of two functions. If $f$ and $g$
are $n$ times differentiable functions, then the product $fg$ is also
differentiable $n$ times and the $n^{\text{th}}$ order derivative is given
by 
\begin{equation}
(fg)^{(n)}=\sum_{k=0}^{n}{\binom{n}{k}}f^{(n-k)}g^{(k)}
\label{eqn_leibniz_rule}
\end{equation}%
with binomial coefficient ${\binom{n}{k}}={\frac{n!}{k!(n-k)!}}$, and $%
f^{(0)}\equiv f$. Here, and throughout the paper, the $n^{\text{th}}$ order
derivative of $f$ is denoted with $\frac{d^{n}}{dt^{n}}f=\mathrm{D}%
^{(r)}f=f^{\left( n\right) }$ as appropriate.

This paper is organized as follows. Section~\ref%
{sec_higher_order_recursive_form} presents the $n^{\text{th}}$-order time
derivatives of the inverse dynamics in recursive form. Section~\ref%
{sec_higher_order_closed_form} presents the $n^{\text{th}}$-order time
derivatives of the EOM in closed form. Section~\ref{sec_results} presents
the application of the proposed derivatives in evaluating third order
inverse dynamics of a robot manipulator and a discussion on its
computational performance. Section~\ref{sec_conclusion} concludes the paper.

\section{Recursive $n$th-Order Inverse Dynamics Algorithm}

\label{sec_higher_order_recursive_form}

\subsection{Kinematics of an open chain}

A body-fixed reference frame (RFR) $\mathcal{F}_{i}$ is attached to link $i$%
. The configuration of body $i$ w.r.t. to a world-fixed inertia frames (IFR)
is represented by a homogenous $4\times 4$ transformation matrix $\mathbf{C}%
_{i}\in SE\left( 3\right) $ \cite{LynchPark2017,Murray,Selig}. The
configuration of body $j$ relative to body $i$ is $\mathbf{C}_{i,j}:=\mathbf{%
C}_{i}^{-1}\mathbf{C}_{j}$. The configuration of body $i$ is given
recursively by the (local) product of exponentials (POE) as $\mathbf{C}%
_{i}=\ \mathbf{C}_{i-1}\mathbf{B}_{i}\exp ({^{i}\mathbf{X}}_{i}q_{i})$.
Therein, ${^{i}\mathbf{X}}_{i}$ is the joint screw coordinate vector in
body-fixed representation, and $\mathbf{B}_{i}$ is the zero reference
configuration of body $i$ and $i-1$. Notice that ${^{i}\mathbf{X}}_{i}$ is
constant.

The twist of link $i$ in \emph{body-fixed} representation is expressed by
the coordinate vector \cite{MUBOScrews2}%
\begin{equation}
\mathbf{V}_{i}=\left( 
\begin{array}{c}
\bm{\omega}%
_{i} \\ 
\mathbf{v}_{i}%
\end{array}%
\right)  \label{DefVb}
\end{equation}%
where $%
\bm{\omega}%
$ denotes the angular velocity of $\mathcal{F}_{i}$ relative to the world
frame $\mathcal{F}_{0}$, and $\mathbf{v}$ is the translational velocity of
the origin of $\mathcal{F}_{i}$ relative to $\mathcal{F}_{0}$, both resolved
in $\mathcal{F}_{i}$. Denote with ${^{i}\mathbf{X}}_{i}$ the screw
coordinate vector of joint $i$ represented in frame $\mathcal{F}_{i}$ at
body $i$. The twist of link $i$, connected to its predecessor link $i-1$ by
joint $i$, is 
\begin{equation}
\mathbf{V}_{i}=\mathbf{Ad}_{\mathbf{C}_{i,i-1}}\mathbf{V}_{i-1}+{^{i}\mathbf{%
X}}_{i}\dot{q}_{i}  \label{Vbrec}
\end{equation}%
with $\mathbf{C}_{i,j}:=\mathbf{C}_{i}^{-1}\mathbf{C}_{j}$, where $\mathbf{Ad%
}_{\mathbf{C}_{i,i-1}}$ is the $6\times 6$ matrix transforming the twist
represented in frame frame $\mathcal{F}_{i-1}$ to its representation in
frame $\mathcal{F}_{i}$, and ${^{i}\mathbf{X}}_{i}\dot{q}_{i}$ is the
relative twist of the two links due to joint $i$.

The recursive relation (\ref{Vbrec}) can be summarized to yield the closed
form relation%
\begin{eqnarray}
\mathbf{V}_{i} &=&\dot{q}_{1}\mathbf{B}_{i,1}+\dot{q}_{2}\mathbf{B}%
_{i,2}+\ldots +\dot{q}_{i}\mathbf{B}_{i,i}  \label{Vb} \\
&=&\mathbf{J}_{i}\left( \mathbf{q}\right) \dot{\mathbf{q}}  \label{Vbi}
\end{eqnarray}%
with the geometric Jacobian of body $i$ in body-fixed representation \cite%
{MUBOScrews2}%
\begin{equation}
\mathbf{J}_{i}\left( \mathbf{q}\right) :=%
\Big%
(\mathbf{B}_{i,1}%
\hspace{-0.5ex}%
\left( \mathbf{q}\right) 
\Big%
|\cdots 
\Big%
|\mathbf{B}_{i,i}%
\hspace{-0.5ex}%
\left( \mathbf{q}\right) 
\Big%
|\mathbf{0}%
\Big%
|\cdots 
\Big%
|\mathbf{0}%
\Big%
)  \label{Jib}
\end{equation}%
where the columns $\mathbf{B}_{i,j}$ are the instantaneous joint screws in
body-fixed representation%
\begin{equation}
\mathbf{B}_{i,j}%
\hspace{-0.5ex}%
\left( \mathbf{q}\right) =\mathbf{Ad}_{\mathbf{C}_{i,j}}\mathbf{X}_{j},\
j\leq i.  \label{Bij}
\end{equation}%
From its construction, follows immediately the recursive relation 
\begin{equation}
\mathbf{B}_{i,j}=\mathbf{Ad}_{\mathbf{C}_{i,i-1}}\mathbf{B}_{i-1,j},j<i,\ \
\ \mathrm{and\ \ \ }\mathbf{B}_{i,i}=\mathbf{X}_{i}.  \label{Brec}
\end{equation}

Notice that $\mathbf{B}_{i,j}$ are usually denoted with $\mathbf{J}_{i,j}$,
but will not be used for sake of simplicity (avoiding the superscript). The
relation%
\begin{equation}
\dot{\mathbf{Ad}}_{\mathbf{C}_{i,i-1}}\mathbf{V}_{i-1}=\dot{q}_{i}\mathbf{ad}%
_{\mathbf{V}_{i}}{^{i}\mathbf{X}}_{i}=-\dot{q}_{i}\mathbf{ad}_{{^{i}\mathbf{X%
}}_{i}}\mathbf{V}_{i}  \label{Addot}
\end{equation}%
yields the recursive expression for the acceleration%
\begin{equation}
\dot{\mathbf{V}}_{i}=\mathbf{Ad}_{\mathbf{C}_{i,i-1}}\dot{\mathbf{V}}_{i-1}+%
\dot{q}_{i}\mathbf{ad}_{\mathbf{V}_{i}}{^{i}\mathbf{X}}_{i}+{^{i}\mathbf{X}}%
_{i}\ddot{q}_{i}.  \label{Vbdotrec}
\end{equation}

\subsection{Newton-Euler Equations of a Rigid Body}

Denote with $\bm{\Theta}_{i}$ the inertia tensor of link $i$ w.r.t. its RFR $%
\mathcal{F}_{i}$, with $m_{i}$ the mass of link $i$, and with ${^{i}}\mathbf{%
d}_{i\text{c}}$ the distance vector from the origin of $\mathcal{F}_{i}$ to
the COM of link $i$. The inertia matrix of body $i$ w.r.t. $\mathcal{F}_{i}$
is then defined as%
\begin{equation}
\mathbf{M}_{i}=\left( 
\begin{array}{cc}
\bm{\Theta}_{i} & {^{i}\widetilde{\mathbf{d}}}_{i\text{c}}m_{i} \\ 
-{^{i}}\widetilde{\mathbf{d}}_{i\text{c}}m_{i} & \mathbf{I}m_{i}%
\end{array}%
\right) .  \label{Mb}
\end{equation}%
The Newton--Euler equations in body-fixed RFR $\mathcal{F}_{i}$ are%
\begin{equation}
\mathbf{W}_{i}=\mathbf{M}_{i}\dot{\mathbf{V}}_{i}-\mathbf{ad}_{\mathbf{V}%
_{i}}^{T}\mathbf{M}_{i}\mathbf{V}_{i}.  \label{NEb}
\end{equation}%
$\mathbf{W}_{i}=(\mathbf{t}_{i},\mathbf{f}_{i})^{T}$ is the wrench applied
to $\mathcal{F}_{i}$ (including gravity), where $\mathbf{t}_{i}\in {\mathbb{R%
}}^{3}$ is the vector of applied torques, and $\mathbf{f}_{i}\in {\mathbb{R}}%
^{3}$ the vector of forces applied at the origin of $\mathcal{F}_{i}$.

\subsection{Higher-Order Forward Kinematics\label{secForwardKin}}

Repeated application of relation (\ref{Vbdotrec}) yields explicit recursive
relations for the jerk, jounce, etc., which were used in \cite{ICRA2017} to
derive a forth-order forward kinematic and a second-order inverse dynamics
algorithm. For higher-order derivatives this becomes rather involved,
however. This can be avoided invoking the relations for the
higher-derivative reported in \cite{Mueller-MMT2019}, which are recursive in
the order of derivative. To this end, introduce%
\begin{equation}
\mathsf{B}_{i,j}(\mathbf{q},\dot{\mathbf{q}}):=\sum_{j<r\leq i}\mathbf{B}%
_{i,r}\left( \mathbf{q}\right) \dot{q}_{r}.  \label{BBij}
\end{equation}%
so that $\mathbf{V}_{i}=\mathsf{B}_{i,0}(\mathbf{q},\dot{\mathbf{q}})$.
Therewith, the $k$th time derivative of the body-fixed twist is determined as%
\begin{equation}
\mathrm{D}^{(k)}\mathbf{V}_{i}=\mathrm{D}^{(k)}\mathsf{B}_{i,0}(\mathbf{q},%
\dot{\mathbf{q}}).
\end{equation}%
The recursive relations (\ref{Brec}) and (\ref{Addot}) give rise to the
recursive expression of time derivative of the joint screws $\mathbf{B}%
_{i,j} $ as%
\begin{equation}
\dot{\mathbf{B}}_{i,j}=\sum_{j<r\leq i}\left[ \mathbf{B}_{i,j},\mathbf{B}%
_{i,r}\right] \dot{q}_{r}=\left[ \mathbf{B}_{i,j},\mathsf{B}_{i,j}\right] =%
\mathbf{ad}_{\mathbf{B}_{i,j}}\mathsf{B}_{i,j}.  \label{Bdot}
\end{equation}%
Higher-order time derivatives of $\mathbf{B}_{i,j}$ are obtained for $k>0$
from (\ref{Bdot}) as%
\begin{eqnarray}
\mathrm{D}^{(k)}\mathbf{B}_{i,j} &=&\sum_{l=0}^{k-1}\tbinom{k-1}{l}[\mathrm{D%
}^{(l)}\mathbf{B}_{i,j},\mathrm{D}^{(k-l-1)}\mathsf{B}_{i,j}]  \label{DkB} \\
&=&\sum_{l=0}^{k-1}\tbinom{k-1}{l}\mathbf{ad}_{\mathrm{D}^{(l)}\mathbf{B}%
_{i,j}}\mathrm{D}^{(k-l-1)}\mathsf{B}_{i,j},\ \ j\leq i.  \nonumber
\end{eqnarray}%
The derivatives of $\mathsf{B}_{i,j}$ follow with (\ref{BBij}) simply as%
\begin{equation}
\mathrm{D}^{(k)}\mathsf{B}_{i,j}=\sum_{j<r\leq i}\sum_{l=0}^{k}\tbinom{k}{l}%
\mathrm{D}^{(l)}\mathbf{B}_{i,r}q_{r}^{(k-l+1)}.  \label{DBk}
\end{equation}%
The important point is that these relations can be easily implemented. The
explicit third- and forth-order relations reported in \cite{ICRA2017} are
special cases, where the recursive relations (\ref{DkB}) and (\ref{DBk}) are
rolled out.

\subsection{Inverse Dynamics}

The backward recursion step in the (standard) inverse dynamics algorithm, in
terms of body-fixed twists and wrenches, \cite{ICRA2017} is 
\begin{equation}
\mathbf{W}_{i}=\mathbf{Ad}_{\mathbf{C}_{i+1,i}}^{T}\mathbf{W}_{i+1}+\mathbf{M%
}_{i}\dot{\mathbf{V}}_{i}-\mathbf{ad}_{\mathbf{V}_{i}}^{T}\mathbf{M}_{i}%
\mathbf{V}_{i}.  \label{back}
\end{equation}%
Higher-order derivatives of the backward recursion step (\ref{back}) follow
with Leibniz' rule (\ref{eqn_leibniz_rule}) as%
\begin{eqnarray}
\mathrm{D}^{(k)}\mathbf{W}_{i} &=&\sum_{r=0}^{k}\tbinom{k}{r}\mathrm{D}^{(r)}%
\mathbf{Ad}_{\mathbf{C}_{i+1,i}}^{T}\mathrm{D}^{(k-r)}\mathbf{W}_{i+1}
\label{higherBack} \\
&&+\mathbf{M}_{i}\mathrm{D}^{(k)}\mathbf{V}_{i}-\sum_{r=0}^{k}\tbinom{k}{r}%
\mathbf{ad}_{\mathrm{D}^{(r)}\mathbf{V}_{i}}^{T}\mathbf{M}_{i}\mathrm{D}%
^{(k-r)}\mathbf{V}_{i}.  \nonumber
\end{eqnarray}%
The relation $\dot{\mathbf{Ad}}_{\mathbf{C}_{i+1,i}}=-\dot{q}_{i+1}\mathbf{ad%
}_{{^{i+1}\mathbf{X}}_{i+1}}\mathbf{Ad}_{\mathbf{C}_{i+1,i}}$ gives rise to 
\begin{equation}
\mathrm{D}^{(k)}\mathbf{Ad}_{\mathbf{C}_{i+1,i}}=-\mathbf{ad}_{{^{i+1}%
\mathbf{X}}_{i+1}}\sum_{r=0}^{k-1}\tbinom{k-1}{r}\mathrm{D}^{(r)}\mathbf{Ad}%
_{\mathbf{C}_{i+1,i}}q_{i+1}^{(k-r)}  \nonumber
\end{equation}%
which allows evaluating (\ref{higherBack}).

\subsection{Higher-Order Inverse Dynamics Algorithm}

\underline{$k$th-Order Forward Kinematics}%
\vspace{-0.5ex}%

\begin{itemize}
\item[$\blacksquare $] Input: $\mathbf{q},\dot{\mathbf{q}},\ddot{\mathbf{q}}%
,\ldots ,\mathbf{q}^{\left( k\right) }$

\item[$\blacksquare $] Preparation run (computation of basic kinematic data) 

\begin{itemize}
\item[$\blacktriangleright $] For body $i=1$:%
\vspace{-1.5ex}%
\begin{eqnarray*}
\mathbf{C}_{1} &=&\mathbf{B}_{1}\exp ({^{1}\mathbf{X}}_{1}q_{1}) \\
\mathbf{B}_{1,1} &=&{^{1}\mathbf{X}}_{1},\ \mathsf{B}_{1,1}=\ {^{1}\mathbf{X}%
}_{1}\dot{q}_{i}
\end{eqnarray*}

\item[$\blacktriangleright $] For $i=2,\ldots ,n$ (recursion over bodies)%
\vspace{-1.5ex}%
\begin{eqnarray*}
\mathbf{C}_{i} &=&\mathbf{C}_{i-1}\mathbf{B}_{i}\exp ({^{i}\mathbf{X}}%
_{i}q_{i}) \\
\mathbf{C}_{i,i-1} &=&\mathbf{C}_{i}^{-1}\mathbf{C}_{i-1} \\
\mathbf{B}_{i,j} &=&\mathbf{Ad}_{\mathbf{C}_{i,i-1}}\mathbf{B}_{i-1,j},\
j=1,\ldots ,i
\end{eqnarray*}

End
\end{itemize}

\item[$\blacksquare $] For $r=0,\ldots ,k$ (recursion over order of
derivative)

\begin{itemize}
\item[$\blacktriangleright $] For body $i=1$:%
\vspace{-1.5ex}
\begin{eqnarray*}
\mathrm{D}^{(r)}\mathbf{V}_{1} &=&{^{1}\mathbf{X}}_{1}q^{\left( r+1\right) }
\\
\mathrm{D}^{(r)}\mathbf{B}_{1,1} &=&\mathbf{0},\ r\geq 1
\end{eqnarray*}

\item[$\blacktriangleright $] For $i=2,\ldots ,n$ (recursion over bodies)%
\vspace{-0.5ex} 
\vspace{-1.5ex}%
\[
\hspace{-2ex}%
\mathrm{D}^{(k)}\mathbf{Ad}_{\mathbf{C}_{i,i-1}}=-\mathbf{ad}_{{^{i}\mathbf{X%
}}_{i}}\sum_{r=0}^{k-1}\tbinom{k-1}{r}\mathrm{D}^{(r)}\mathbf{Ad}_{\mathbf{C}%
_{i,i-1}}q_{i}^{(k-r)}
\]

\begin{itemize}
\item[$\bullet $] For $j=i,i-1.\ldots ,1$%
\begin{align*}
\hspace{-2ex}%
\mathrm{D}^{(r)}\mathbf{B}_{i,j}& =\sum_{l=0}^{r-1}\tbinom{r-1}{l}\mathbf{ad}%
_{\mathrm{D}^{(l)}\mathbf{B}_{i,j}}\mathrm{D}^{(r-l-1)}\mathsf{B}_{i,j},r>1
\\
\hspace{-2ex}%
\mathrm{D}^{(r)}\mathsf{B}_{i,j}& =\sum_{j\leq k\leq i}\sum_{l=0}^{r}\tbinom{%
r}{l}\mathrm{D}^{(l)}\mathbf{B}_{i,r}q_{j}^{(r-l+1)} \\
\hspace{-2ex}%
\mathrm{D}^{(r)}\mathbf{V}_{i}& =\mathrm{D}^{(r)}\mathsf{B}_{i,1}(\mathbf{q},%
\dot{\mathbf{q}})
\end{align*}%
end
\end{itemize}

end
\end{itemize}

end

\item[$\blacksquare $] Output: $\mathbf{C}_{i},\mathrm{D}^{(k)}\mathbf{V}%
_{i},\mathrm{D}^{(k)}\mathbf{B}_{i,j},\mathrm{D}^{(k)}\mathsf{B}_{i,j}$
\end{itemize}

\underline{$k$th-Order Inverse Dynamics}%
\vspace{-0.5ex}%

\begin{itemize}
\item[$\blacksquare $] Input: $\mathbf{C}_{i},\mathrm{D}^{(k)}\mathbf{V}%
_{i}^{\mathrm{b}}$

\item[$\blacksquare $] For $i=n-1,\ldots ,1$ 
\vspace{-3ex}%
\end{itemize}

\begin{eqnarray*}
\mathrm{D}^{(k)}\mathbf{W}_{i} &=&\sum_{r=0}^{k}\tbinom{k}{r}\mathrm{D}^{(r)}%
\mathbf{Ad}_{\mathbf{C}_{i+1,i}}^{T}\mathrm{D}^{(k-r)}\mathbf{W}_{i+1} \\
&&+\mathbf{M}_{i}^{\text{b}}\mathrm{D}^{(k)}\mathbf{V}_{i} \\
&&-\sum_{r=0}^{k}\tbinom{k}{r}\mathbf{ad}_{\mathrm{D}^{(r)}\mathbf{V}%
_{i}}^{T}\mathbf{M}_{i}\mathrm{D}^{(k-r)}\mathbf{V}_{i} \\
\mathrm{D}^{(k)}Q_{i}%
\hspace{-1.6ex}
&=&%
\hspace{-1.6ex}%
{^{i}\mathbf{X}}_{i}^{T}\mathrm{D}^{(k)}\mathbf{W}_{i}
\end{eqnarray*}%
end

\begin{itemize}
\item[$\blacksquare $] Output: $\mathrm{D}^{(k)}\mathbf{Q}$%
\vspace{2ex}%
\end{itemize}


\section{$n$th-Order Time Derivatives of Equations of Motion}

\label{sec_higher_order_closed_form}

\subsection{EOM in Closed Form}

The individual twists of all bodies are summarized in the vector $\mathsf{V}%
\in {\mathbb{R}}^{6n}$, which is referred to as the system twist in
body-fixed representation. It is determined as%
\begin{equation}
\mathsf{V}=\mathsf{J}\dot{\mathbf{q}}  \label{Vsys}
\end{equation}%
with the system Jacobian $\mathsf{J}\left( \mathbf{q}\right) $. The latter
admits the factorization%
\begin{equation}
\mathsf{J}=\mathsf{AX}  \label{JbSys}
\end{equation}%
in terms of the block-triangular and block-diagonal matrices 
\begin{eqnarray}
\mathsf{A}\left( \mathbf{q}\right)  &=&\left( 
\begin{array}{ccccc}
\mathbf{I} & \mathbf{0} & \mathbf{0} &  & \mathbf{0} \\ 
\mathbf{Ad}_{\mathbf{C}_{2,1}} & \mathbf{I} & \mathbf{0} & \cdots  & \mathbf{%
0} \\ 
\mathbf{Ad}_{\mathbf{C}_{3,1}} & \mathbf{Ad}_{\mathbf{C}_{3,2}} & \mathbf{I}
&  & \mathbf{0} \\ 
\vdots  & \vdots  & \ddots  & \ddots  &  \\ 
\mathbf{Ad}_{\mathbf{C}_{n,1}} & \mathbf{Ad}_{\mathbf{C}_{n,2}} & \cdots  & 
\mathbf{Ad}_{\mathbf{C}_{n,n-1}} & \mathbf{I}%
\end{array}%
\right) \hspace{-1.5ex} \\
\mathsf{X} &=&\left( 
\begin{array}{ccccc}
{^{1}\mathbf{X}}_{1} & \mathbf{0} & \mathbf{0} &  & \mathbf{0} \\ 
\mathbf{0} & {^{2}\mathbf{X}}_{2} & \mathbf{0} & \cdots  & \mathbf{0} \\ 
\mathbf{0} & \mathbf{0} & {^{3}\mathbf{X}}_{3} &  & \mathbf{0} \\ 
\vdots  & \vdots  & \ddots  & \ddots  &  \\ 
\mathbf{0} & \mathbf{0} & \cdots  & \mathbf{0} & {^{n}}\mathbf{X}_{n}%
\end{array}%
\right)   \nonumber  \label{Ab}
\end{eqnarray}%
where ${^{i}\mathbf{X}}_{i}$ is the screw coordinate vector associated to
joint $i$ represented in the body-frame of body $i$. The vectors ${^{i}%
\mathbf{X}}_{i}$ are constant due to the body-fixed representation. The
matrix $\mathbf{Ad}_{\mathbf{C}_{i,j}}$ transforms screw coordinates
represented in the reference frame at body $j$ to those represented in the
frame on body $i$ \cite{LynchPark2017,Murray,Selig}. A central relation for
deriving the EOM in closed form is the following expression for the time
derivative of the matrix $\mathsf{A}$ and thus of the system Jacobian \cite%
{MUBOScrews2}%
\begin{equation}
\dot{\mathsf{J}}\left( \mathbf{q},\dot{\mathbf{q}}\right) =-\mathsf{A}\left( 
\mathbf{q}\right) \mathsf{a}\left( \dot{\mathbf{q}}\right) \mathsf{J}\left( 
\mathbf{q}\right)   \label{Jdot}
\end{equation}%
where 
\begin{equation}
\mathsf{a}\left( \dot{\mathbf{q}}\right) =\mathrm{diag}~(\dot{q}_{1}\mathbf{%
ad}_{{{^{1}}\mathbf{X}_{1}}},\ldots ,\dot{q}_{n}\mathbf{ad}_{{{^{n}}\mathbf{X%
}_{n}}}).  \label{a}
\end{equation}%
This gives rise to the closed form expressions for the system acceleration%
\begin{equation}
\dot{\mathsf{V}}=\mathsf{J}\ddot{\mathbf{q}}-\mathsf{AaJ}\dot{\mathbf{q}}=%
\mathsf{J}\ddot{\mathbf{q}}-\mathsf{AaV}.  \label{VbdotMat}
\end{equation}%
For calculating the derivatives, the time derivative of matrix $\mathsf{A}$
will be needed. It can be shown that the derivative of $\mathsf{A}$ is \cite%
{MUBOScrews2}%
\begin{equation}
\dot{\mathsf{A}}\left( \mathbf{q},\dot{\mathbf{q}}\right) =\mathsf{A}\left( 
\mathbf{q}\right) \mathsf{a}-\mathsf{A}\left( \mathbf{q}\right) \mathsf{a}%
\left( \dot{\mathbf{q}}\right) \mathsf{A}\left( \mathbf{q}\right) .
\label{Adot}
\end{equation}%
Clearly, the derivative (\ref{Jdot}) of the system Jacobian is recovered as $%
\dot{\mathsf{J}}=\dot{\mathsf{A}}\mathsf{X}$ noting that $\mathsf{aX}\equiv 
\mathbf{0}$.

The generalized mass and Coriolis matrix in the EOM (\ref{EOM1}) of a simple
kinematic chain mounted at the ground are found via Jourdain's principle of
virtual power as (or likewise as the Lagrange equations) \cite{MUBOScrews2}%
\begin{equation}
\mathbf{M}\left( \mathbf{q}\right) =\mathsf{J}^{T}\mathsf{MJ,\ \ \ \ }%
\mathbf{C}\left( \mathbf{q},\dot{\mathbf{q}}\right) =\mathsf{J}^{T}\mathsf{CJ%
}  \label{MC}
\end{equation}%
where%
\begin{eqnarray}
\mathsf{M} &:=&\mathrm{diag}\,(\mathbf{M}_{1},\ldots ,\mathbf{M}_{n})
\label{Msys} \\
\mathsf{C}(\mathbf{q},\dot{\mathbf{q}},\mathsf{V}\left( \dot{\mathbf{q}}%
\right) ) &:=&-\mathsf{MAa}-\mathsf{b}^{T}\mathsf{M}.  \label{Csys}
\end{eqnarray}%
Therein, $\mathbf{M}_{i}$ is the (constant) $6\times 6$ inertia matrix of
body $i$ expressed in the body-frame, and 
\begin{equation}
\mathsf{b}\left( \mathsf{V}\right) :=\mathrm{diag}~(\mathbf{ad}_{\mathbf{V}%
_{1}},\ldots ,\mathbf{ad}_{\mathbf{V}_{n}}).  \label{b}
\end{equation}%
A closed form of the EOM is obtained after replacing the system twist by (%
\ref{Vsys}). Alternatively, first the kinematic relation (\ref{Vsys}) and
then the coefficient matrices in (\ref{Csys}) are evaluated for a given
state $\mathbf{q},\dot{\mathbf{q}}$. The generalized gravity forces are
given as%
\begin{equation}
\mathbf{Q}_{\mathrm{grav}}\left( \mathbf{q}\right) =\mathsf{J}^{T}\mathsf{MU}%
\dot{\mathbf{V}}_{0}
\end{equation}%
with%
\begin{equation}
\dot{\mathbf{V}}_{0}=\left( 
\begin{array}{c}
\mathbf{0} \\ 
^{0}\mathbf{g}%
\end{array}%
\right),\mathsf{U}\left( \mathbf{q}\right)=\mathsf{A}\left( 
\begin{array}{c}
\mathbf{I} \\ 
\mathbf{0} \\ 
\vdots \\ 
\mathbf{0}%
\end{array}%
\right)=\left( 
\begin{array}{c}
\mathbf{Ad}_{\mathbf{C}_{1}}^{-1} \\ 
\mathbf{Ad}_{\mathbf{C}_{2}}^{-1} \\ 
\vdots \\ 
\mathbf{Ad}_{\mathbf{C}_{n}}^{-1}%
\end{array}%
\right).  \label{eq_U}
\end{equation}%
Here, $^{0}\mathbf{g}$ is the vector of gravitational acceleration expressed
in the inertial frame, which is transformed to the individual bodies by $%
\mathsf{U}$.

\subsection{Higher-Order Time Derivatives of the EOM}

In this section, the expressions for $n^{\text{th}}$ order time derivatives
of the EOM in closed form are derived. Applying Leibniz's rule (\ref%
{eqn_leibniz_rule}) on the EOM (\ref{EOM1}), one gets 
\begin{eqnarray}
\mathbf{Q}^{(n)} &=&\sum_{k=0}^{n}{\binom{n}{k}}\mathbf{M}^{(n-k)}\ddot{%
\mathbf{q}}^{(k)}+  \label{EOMn} \\
&&\sum_{k=0}^{n}{\binom{n}{k}}\mathbf{C}^{(n-k)}\dot{\mathbf{q}}^{(k)}+%
\mathbf{Q}_{\mathrm{grav}}^{(n)}  \nonumber
\end{eqnarray}%
which in an expanded form can also be written as 
\begin{align*}
\mathbf{Q}^{(n)}-\mathbf{Q}_{\mathrm{grav}}^{(n)}& =\mathbf{M}\ddot{\mathbf{q%
}}^{(n)}+(n\mathbf{M}^{(1)}+\mathbf{C})\ddot{\mathbf{q}}^{(n-1)}+\ldots \\
& +\left[ {\binom{n}{k}}\mathbf{M}^{(n-k)}+{\binom{n}{k+1}}\mathbf{C}%
^{(n-k-1)}\right] \ddot{\mathbf{q}}^{(k)} \\
& +\left[ {\binom{n}{k-1}}\mathbf{M}^{(n-k+1)}+{\binom{n}{k}}\mathbf{C}%
^{(n-k)}\right] \dot{\mathbf{q}}^{(k)} \\
& +\ldots +(\mathbf{M}^{(n)}+n\mathbf{C}^{(n-1)})\ddot{\mathbf{q}}+\mathbf{C}%
^{(n)}\dot{\mathbf{q}}.
\end{align*}%
One can arrive at first order and second order time derivative of EOM by
substituting $n=1$ and $n=2$ in (\ref{EOMn}) respectively 
\begin{align}
\hspace{-2ex}\dot{\mathbf{Q}} &=\mathbf{M}\dddot{\mathbf{q}}+(\dot{\mathbf{M}%
}+\mathbf{C})\ddot{\mathbf{q}}+\dot{\mathbf{C}}\dot{\mathbf{q}}+\dot{\mathbf{%
Q}}_{\mathrm{grav}}\hspace{-0.5ex}  \label{EOM1stOrder} \\
\hspace{-2ex}\ddot{\mathbf{Q}} &=\mathbf{M}\ddot{\ddot{\mathbf{q}}}+(2\dot{%
\mathbf{M}}+\mathbf{C})\dddot{\mathbf{q}}+(\ddot{\mathbf{M}}+2\dot{\mathbf{C}%
})\ddot{\mathbf{q}}+\ddot{\mathbf{C}}\dot{\mathbf{q}}+\ddot{\mathbf{Q}}_{%
\text{grav}}\hspace{-0.5ex}  \label{EOM2ndOrder}
\end{align}%
The matrix coefficient $\mathbf{P}_k$ of $k^{\text{th}}$ derivative of $%
\mathbf{q}$ in the $n^{\text{th}}$ order time derivative of the EOM
expressed as 
\begin{eqnarray}
\label{nthorderQ}
\mathbf{Q}^{(n)} - \mathbf{Q}_{\mathrm{grav}}^{(n)} = \mathbf{M}\mathbf{q}%
^{(n+2)} + \ldots + \mathbf{P}_k\mathbf{q}^{(k)} + \\ \notag
\ldots + \mathbf{C}^{(n)} \mathbf{q}^{(1)}  
\end{eqnarray}%
is given by 
\begin{equation}
\mathbf{P}_k = {\binom{n }{k-2}}\mathbf{M}^{(n-k+2)} + {\binom{n }{k-1}}%
\mathbf{C}^{(n-k+1)}  \label{EOMn_coeff}
\end{equation}%
for all $2 \leq k \leq n+1, n \geq 1$.
For example, the matrix coefficient of $\mathbf{q}^{(3)}$ in $%
2^{\text{nd}}$ order time derivatives of the EOM can be computed by
substituting $k=3,n=2$ in (\ref{EOMn_coeff}) as $(2\dot{\mathbf{M}}+\mathbf{C%
})$ which can be verified from (\ref{EOM2ndOrder}). In the following, higher
order derivatives of various kinematic and dynamic quantities are presented
which are needed in evaluating higher order derivatives of the generalized
forces in (\ref{nthorderQ}).


\subsubsection{Higher Order Kinematics}

For all $n \geq 1$, the $n^{\text{th}}$ order time derivative of the matrix $%
\mathsf{A}$ can be expressed as 
\begin{eqnarray}  \label{nthorderA}
\mathsf{A}^{(n)} = \sum _{k=0}^{n-1}{\binom{n-1 }{k}} \mathsf{A}^{(n-1-k)} 
\mathsf{a}^{(k)} - \\
\sum _{k=0}^{n-1}{\binom{n-1 }{k}} \mathsf{A}^{(n-1-k)} \sum _{j=0}^{k}{%
\binom{k }{j}} \mathsf{a}^{(k-j)} \mathsf{A}^{(j)}.  \nonumber
\end{eqnarray}
which requires the higher order derivatives of the matrix $\mathsf{a}$. The $%
n^{\text{th}}$ order time derivative of the matrix $\mathsf{a}$ is given by 
\begin{equation}  \label{nthordera}
\mathsf{a}^{\left(n\right)}\left(\mathbf{q}^{\left(n\right)}\right) =\mathrm{%
diag}~(q^{\left(n\right)}_{1}\mathbf{ad}_{{{^{1}}\mathbf{X}_{1}}},\ldots
,q^{\left(n\right)}_{n}\mathbf{ad}_{{{^{n}}\mathbf{X}_{n}}}).
\end{equation}%
By performing a book keeping of all the previous derivatives of $\mathsf{A}$
and $\mathsf{a}$, $n^{\text{th}}$ derivative of the matrix $\mathsf{A}$ can
be easily computed. %
%

Using the (\ref{nthorderA}) and (\ref{JbSys}), $n^{\text{th}}$ time
derivative of the system level Jacobian matrix can be computed as 
\begin{equation}
\mathsf{J}^{(n)}=\mathsf{A}^{(n)}\mathsf{X}.  \label{nthorderJacobian}
\end{equation}%
Similarly, $n^{\text{th}}$ time derivative of the system velocities can be
obtained via 
\begin{equation}
\mathsf{V}^{(n)}=\sum_{k=0}^{n}{\binom{n}{k}}\mathsf{J}^{(n-k)}\dot{\mathbf{q%
}}^{(k)}.
\end{equation}

\subsubsection{Higher Order Mass-Inertia matrix}

The $n^{\text{th}}$ time derivative of the mass-inertia matrix $\mathbf{M}$
is given by 
\begin{equation}
\mathbf{M}^{(n)}=\sum_{k=0}^{n}{\binom{n}{k}}{\mathsf{J}}^{(n-k)T}\mathsf{M}{%
\mathsf{J}}^{(k)}
\end{equation}%
which could be readily computed using higher order derivatives of the system
level Jacobian (\ref{nthorderJacobian}).

\subsubsection{Higher Order Coriolis-Centrifugal Matrix}


The $n^{\text{th}}$ time derivative of the Coriolis-Centrifugal matrix $%
\mathbf{C}$ is given by: 
\begin{equation}
\mathbf{C}^{(n)}=\sum_{k=0}^{n}{\binom{n}{k}}{\mathsf{J}}^{(n-k)T}%
\sum_{j=0}^{k}{\binom{k}{j}}\mathsf{C}^{(k-j)}{\mathsf{J}}^{(j)}
\label{nthorderCoriolis}
\end{equation}%
which necessities the higher order derivatives of $\mathsf{C}$. The
expression of $n^{\text{th}}$ time derivative of system level
Coriolis-Centrifugal matrix is given by 
\begin{equation}
\mathsf{C}^{(n)}=-\mathsf{M}\sum_{k=0}^{n}{\binom{n}{k}}{\mathsf{A}}^{(n-k)}{%
\mathsf{a}}^{(k)}-\mathsf{b}^{(n)T}{\mathsf{M}}
\label{nthorderSystemCoriolis}
\end{equation}%
which requires the $n^{\text{th}}$ order derivative of the matrix $\mathsf{b}
$ computed as 
\begin{equation}
\mathsf{b}^{\left( n\right) }\left( \mathsf{V}^{\left( n\right) }\right) :=%
\mathrm{diag}~\left( \mathbf{ad}_{\mathbf{V}_{1}^{\left( n\right) }},\ldots ,%
\mathbf{ad}_{\mathbf{V}_{n}^{\left( n\right) }}\right)   \label{nthorderb}
\end{equation}%
and higher order derivatives of $\mathsf{A}$ and $\mathsf{a}$ available in (%
\ref{nthorderA}) and (\ref{nthordera}) respectively.

\subsubsection{Higher Order Gravity Force Vector}

The $n^{\text{th}}$ order time derivative of the vector of gravity forces $%
\mathbf{Q}_{\text{grav}}$ is given by 
\begin{equation}
\mathbf{Q}_{\text{grav}}^{(n)} = \sum_{k=0}^{n}{\binom{n}{k}}{\mathsf{J}}%
^{(n-k)T} \mathsf{M} \mathsf{A}^{\left( k\right) } \left( 
\begin{array}{c}
\mathbf{I} \\ 
\mathbf{0} \\ 
\vdots \\ 
\mathbf{0}%
\end{array}%
\right) \dot{\mathbf{V}}_{0}.
\end{equation}%
%
%
%
%
%
%
%

\section{Results and Discussion}

\label{sec_results} The $n^{\text{th}}$ order recursive inverse dynamics
algorithm presented in Section~\ref{sec_higher_order_recursive_form} and
higher order closed form equations of motion presented in Section~\ref%
{sec_higher_order_closed_form} were implemented in MATLAB\footnote{%
The source code as well as robot data will be made publicly available after
the acceptance of the paper.}. This section presents their application to
the computation of higher order inverse dynamics of a robot manipulator and
presents a discussion on their computational efficiency.

\subsection{Example: Third Order Inverse Dynamics}

\begin{figure}[!htb]
\centering
\includegraphics[width=0.2\linewidth]{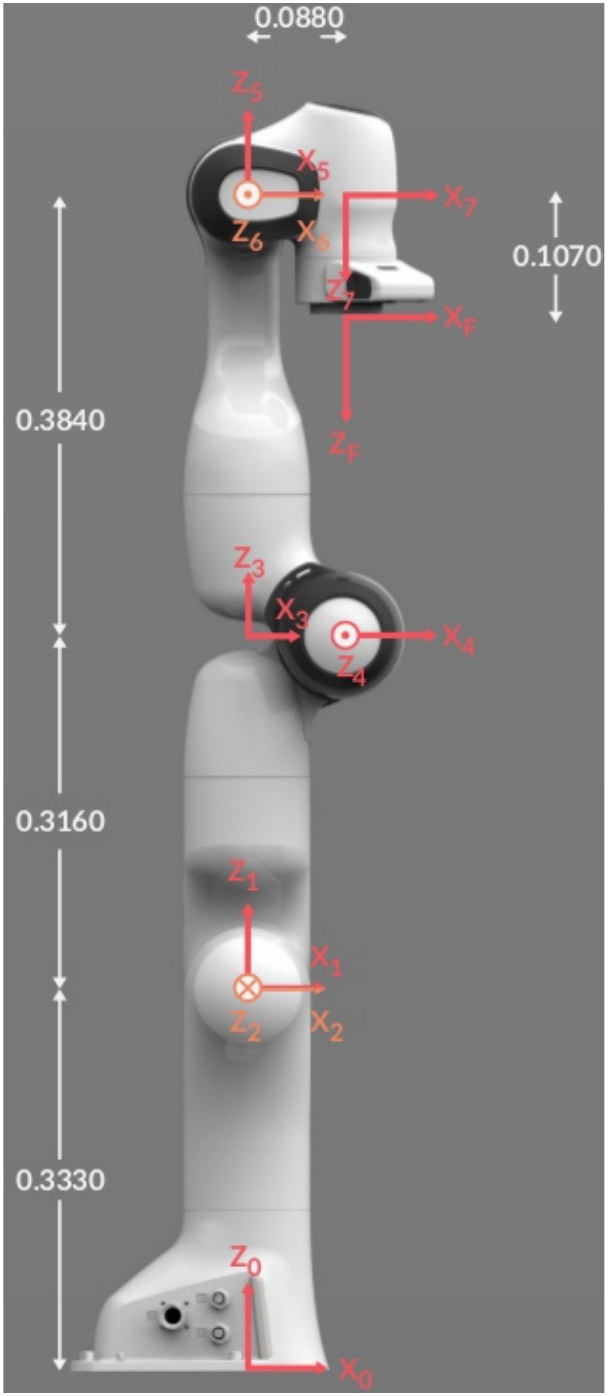}
\hfill \includegraphics[width=0.67\linewidth]{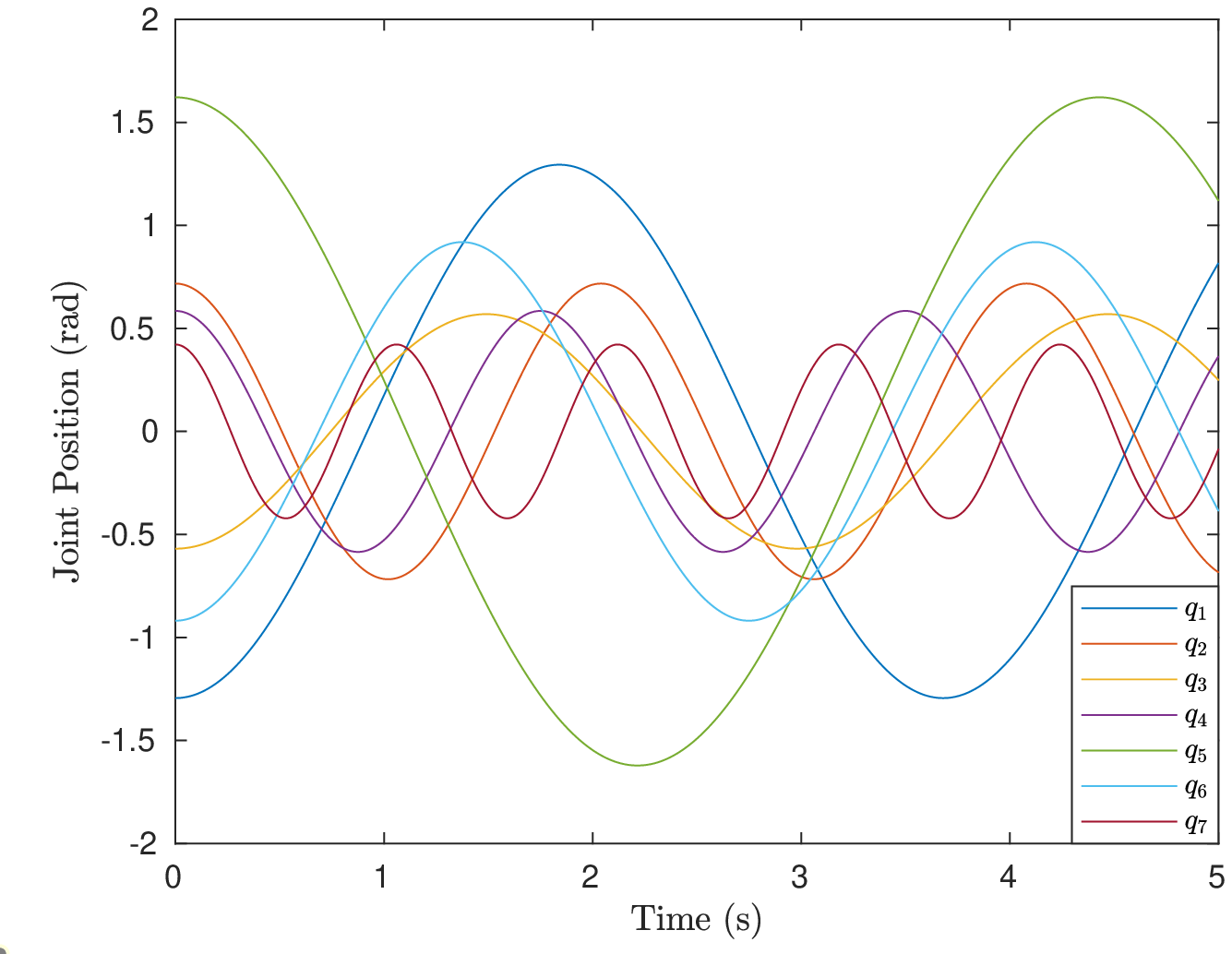} \vspace{-2ex}
\caption{Franka Emika Panda robot with geometric parameters in the zero
configuration (i.e. $\mathbf{q} = 0$) (left) and input joint motion
trajectory (right)}
\label{figFranka}
\end{figure}

\begin{figure}[!htb]
\centering
\includegraphics[width=\linewidth]{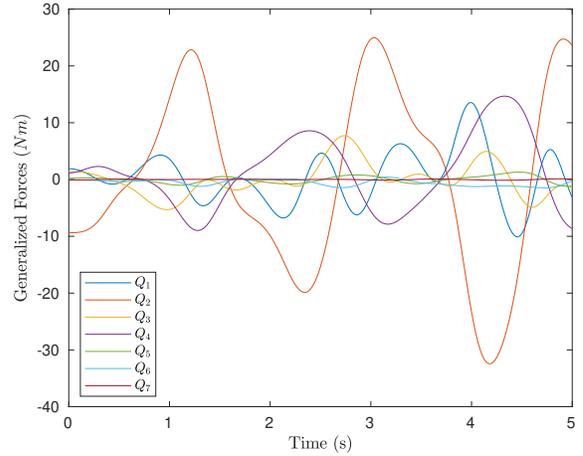}
\caption{Inverse Dynamics}
\label{figQ}
\end{figure}

\begin{figure}[!htb]
\vspace{-2ex} \centering
\includegraphics[width=\linewidth]{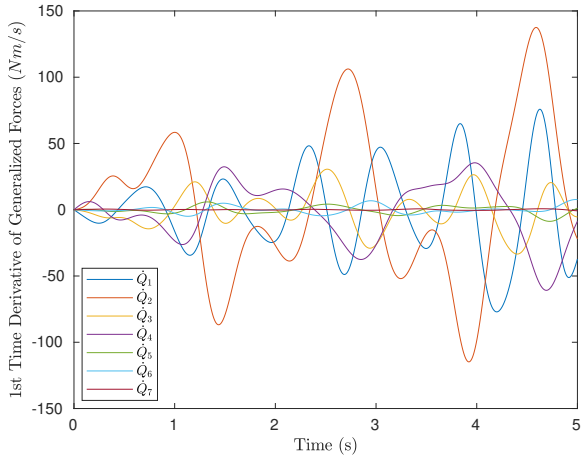} \vspace{-4ex}
\caption{1st Order Inverse Dynamics}
\label{figQd}
\end{figure}

\begin{figure}[!htb]
\centering
\includegraphics[width=\linewidth]{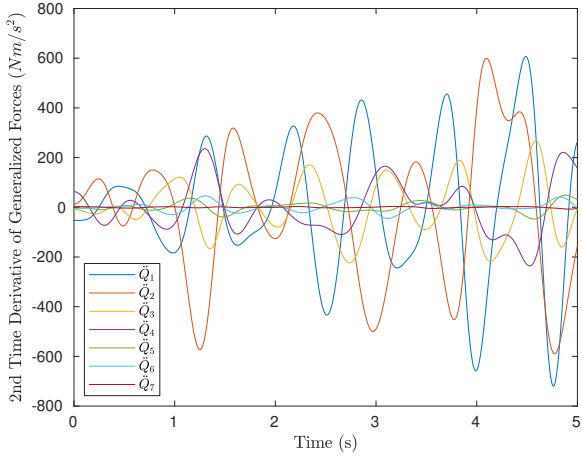} \vspace{-4ex}
\caption{2nd Order Inverse Dynamics}
\label{figQ2d}
\end{figure}

\begin{figure}[!htb]
\centering
\includegraphics[width=\linewidth]{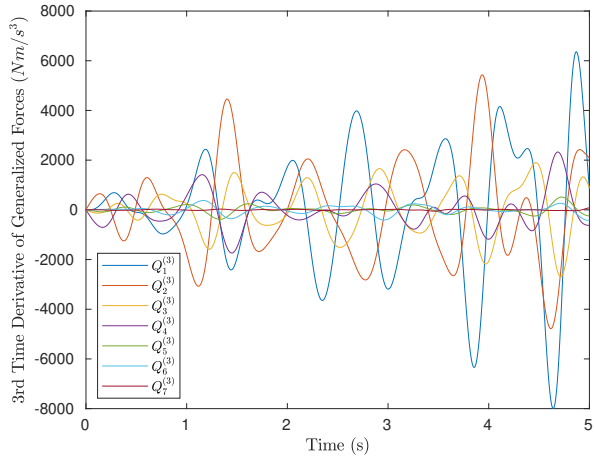} \vspace{-4ex}
\caption{3rd Order Inverse Dynamics}
\label{figQ3d}
\end{figure}

Both algorithms are applied to compute the third order inverse dynamics of a
six degrees of freedom (DOF) Franka Emika Panda robot as shown in Figure~\ref%
{figFranka} (left). Using the geometric information provided in Figure~\ref%
{figFranka}, it is straight-forward to deduce the joint screw coordinate
vectors $^{i}\mathbf{X}_{i}$ and the relative reference configuration of all
the links $\mathbf{B}_{i}$ thanks to the simplicity of this modeling
approach as compared to the DH parameters (for details see~\cite{MUBOScrews1}%
). The mass--inertia data w.r.t. the RFRs was determined as reported in~\cite%
{GazPanda2019} providing the body-fixed link mass matrices (\ref{Mb}). Due
to space the limitation, details are omitted here. The joint trajectory,
taking from \cite{GazPanda2019}, as shown in Figure~\ref{figFranka} (right)
is used as the input motion trajectory for the system and the inverse
dynamics of the manipulator arm is computed up to $3^{\text{rd}}$ order.
Figure~\ref{figQ}, \ref{figQd}, \ref{figQ2d} and \ref{figQ3d} show the
higher order inverse dynamics results (i.e. $\mathbf{Q}(t),\dot{\mathbf{Q}}%
(t),\ddot{\mathbf{Q}}(t),\dddot{\mathbf{Q}}(t)$). The results were verified
with numerical differentiation of the generalized forces. Both recursive and
closed form algorithms indeed yield the same solution of the higher-order
forward kinematics and inverse dynamics problem (within the numerical
accuracy).

\vspace{-2ex}

\vspace{-2ex} 


\subsection{Discussion on Computational Performance}

The computational performance of the recursive and closed form algorithms
was evaluated by measuring the total CPU time spent in $10000$ evaluations
of $2^{\text{nd}}$ order inverse dynamics on a standard laptop with Intel
Core i7-6600U CPU clocked at 2.6 GHz and 16 GB RAM. It was found that $n^{%
\text{th}}$ order recursive algorithm takes $117$ seconds and the $n^{\text{%
th}}$ order closed form algorithm takes a total of $124$ seconds for $10000$
calls. Hence, it can be noticed that the recursive version of the algorithm
slightly outperforms the closed form version of the algorithm as expected.
It is to be noted that the closed form expressions were implemented using
usual matrices in MATLAB and computational performance can be improved by
exploiting the sparse matrix algebra. In both cases, the equations were
implemented without any optimization, e.g. avoiding multiplication with
zeros etc. These computation times are only preliminary indicators and will
reduce in an optimized C++ implementation.

Additionally, we investigated how the computational efficiency of these
general purpose algorithms for $n^{\text{th}}$ order derivatives compares to
the hand--crafted recursive~\cite{ICRA2017} and closed form algorithms~\cite%
{MuellerKumarMUBO2020} for $2^{\text{nd}}$ order inverse dynamics based on
our previous work. It was found that these hand--crafted algorithms for $2^{%
\text{nd}}$ inverse dynamics are faster than $n^{\text{th}}$ order
algorithms with $n=2$. For example, the recursive $2^{\text{nd}}$ order
inverse dynamics algorithm requires only $24$ seconds and the corresponding
closed form version requires $69$ seconds for $10000$ evaluations.
Apparently, the generality of these algorithms comes at some computational
expense which we believe is justified for the benefit that they admit
evaluating derivative of any order without the need to derive them by hand
or with automatic differentiation tools. This seems even justified for
specific applications, where one would use a hand--crafted algorithm, given
the low runtime and the potential for performance improvement when
efficiently implemented. The presented algorithms can also complement and
improve the functionality of rigid body dynamics libraries.


\section{Conclusion}

\label{sec_conclusion} This paper presents novel recursive and closed form
expressions for the $n^{\text{th}}$ order time derivative of the EOM of a
kinematic chain. Building upon the Lie formulation of the EOM the
formulations are advantageous as they are expressed in terms of joint screw
coordinates, and thus facilitate parameterization in terms of vector
quantities that can be easily obtained. With these relations, general
geometric formulations for the $n^{\text{th}}$ order time derivatives as
needed for motion planning and control are now available. Future research
will also address the $n^{\text{th}}$ order time derivatives of general
mechanisms with kinematic loops and an efficient C++ based implementation in
Hybrid Robot Dynamics (HyRoDyn) software framework~\cite{KumarHyRoDyn2020}.




%

\section*{ACKNOWLEDGMENT}

This work has been performed in the VeryHuman project funded by the German
Aerospace Center (DLR) with federal funds (Grant Number: FKZ 01IW20004) from
the Federal Ministry of Education and Research (BMBF). The second author
acknowledges the support of the LCM K2 Center for Symbiotic Mechatronics
within the framework of the Austrian COMET-K2 program.


\end{document}